# Shift-Memory Network for Temporal Scene Segmentation

Guo Cheng
Indiana University – Purdue University Indianapolis
Indianapolis IN 46202 USA
guocheng@iu.edu

Jiang Yu Zheng
Indiana University – Purdue University Indianapolis
Indianapolis IN 46202 USA
jzheng@iupui.edu

## Abstract

*Semantic segmentation has achieved great accuracy in understanding spatial layout. For real-time tasks based on dynamic scenes, we extend semantic segmentation in temporal domain to enhance the spatial accuracy with motion. We utilize a shift-mode network over streaming input to ensure zero-latency output. For the data overlap under shifting network, this paper identifies repeated computation in fixed periods across network layers. To avoid this redundancy, we derive a Shift-Memory Network (SMN) from encoding-decoding baseline to reuse the network values without accuracy loss. Trained in patch-mode, the SMN extracts the network parameters for SMN to perform inference promptly in compact memory. We segment dynamic scenes from 1D scanning input and 2D video. The experiments of SMN achieve equivalent accuracy as shift-mode but in faster inference speeds and much smaller memory. This will facilitate semantic segmentation in real-time application on edge devices.*

## 1. Introduction

The CNN based deep learning has achieved remarkable success in computer vision tasks such as object detection and semantic segmentation. The hierarchical network structure shows high performance in pattern recognition done with patches, e.g., bounding-boxes or windows. This work extends the semantic segmentation to spatial-temporal domain such as a video clip to enhance the accuracy over a single image, which has motion for control tasks such as autonomous driving. For dynamic scenes in streaming data, video volume (3D) semantic segmentation performed in offline patch-mode has a long latency in responding to input, which makes it difficult for real-time applications such as autonomous driving [1,57]. A video with motion traces adds temporal associations to improve the segmentation accuracy [30-46], but it requires additional computation and much more memory, which is a challenge to apply it on edge devices.

Instead of using patch-mode in semantic segmentation, shift-mode moves CNN module along time axis consecutively with a small shift for fast responding. The minimum latency is achieved when shifting unit is one. It can generate output at time t based on previous inputs $I(x,t)$, where $x \in \Re^2$ is spatial data and $t \in \Re^1$ is time. The latest output thus is connected to a history with continuity and causality. The reference to the past based on Markov principle can improve the classification at the latest moment [64] for decision making and machine control. However, a small shifting unit of network leads to a large network overlap on the input data. As illustrated in Fig.1, sequential semantic segmentation turns the traditional encoder-decoder module [10] from spatial domain to temporal domain. Patch-mode has no data overlap but long latency, and shift-mode reduces latency but results in large data overlap and computing redundancy.

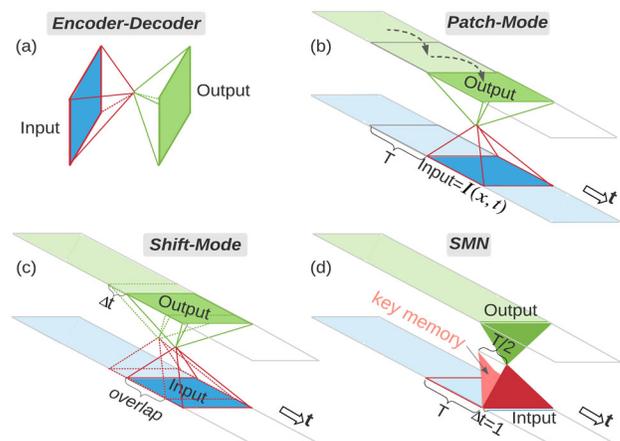

Figure 1: Semantic segmentation applied to the temporal domain on streaming 1D input. (a) Image semantic segmentation with encoder and decoder modules. (b) Patch-mode jump of network on sequential input has long output latency. (c) Naïve shift-mode outputs every moment for real-time input but involves data overlaps and redundant computation. (d) By exploiting the repeated computation in network hierarchy during shifting, our SMN achieves much faster inferencing with less memory.

In this work, we apply semantic segmentation in temporal domain sequentially to include long-span motion and temporal association into account. To simultaneously output and avoid repetitive computation in shifting, we construct a new inference network named Shift-Memory Network (SMN) to reuse the network data computed in previous time at multi-levels. Outlined in Fig. 1(d), SMN is derived from naïve shift-mode of encoder-decoder module but has no repeated computation between consecutive windows. Specifically, in each neural layer of encoding, SMN precisely links the temporal moments in the history to the latest input according to the repeating periods identified in the network hierarchy. Therefore, SMN achieves a fast responding time to produce consistent motion trace from each input, with only twice of computation than a single 2D image. We test the SMN on 1D×t data scanned from driving scenes [42,53] and 2D video including DAVIS [54,55] and Highway Driving Dataset [56], respectively.

The main contributions of this work: (i) Employing semantic segmentation to the spatial-temporal streaming data; our video baseline has richer context for temporal inference in understanding scenes and dynamic events. (ii) We formulate the period of network redundancy in shift-mode, which provides an intrinsic view of CNN hierarchy in sequential semantic segmentation. (iii) Trained in patch-mode with a large memory, our proposed SMN inference achieves the equivalent accuracy as the trained network on video but uses the minimum memory in fast inferencing.

## 2. Related Work

### 2.1. 2D Segmentation on Images

Along with successful CNN of deep learning in object detection [1-8], semantic segmentation has classified pixel-wise spatial regions [9,10] for intelligent vehicle, indoor scenes, and surveillance. Various network updates exploit spatial association in images to improve accuracy [11-20]. Some new structure has increased frame rate faster than video rate [65,66].

However, such an image segmentation does not involve time association and missed motion information. With temporal data, the accuracy will be increased in a certain degree as in the following, though it requires more computation time than image segmentation.

### 2.2. 2D + t Segmentation for Videos

To enhance spatial recognition, temporal continuity among video frames has been shared at latent levels of deep network. Due to a larger memory involved, such efforts are applied at high latent levels. Some techniques connect ad-hoc links in temporal connection.

**LSTM**: RNN [21,22] or LSTM [23] embedded CNN models have been commonly used in semantic segmentation and video recognition [24-32,61].

**Attention**: methods adopt attention technique [33] to query temporal relation from past frames [34-36,60] by matching regions in spatial layout. The costly computation makes them difficult for full-view video semantic segmentation in real-time.

**Shift**: To avoid intensive computation in video volume, some methods shift part of channels in 2D-CNN structure in temporal domain [37-41]. Only a few frames before and after are associated. Such short span "motion blur" in hidden layers may not fully gain long motion in video.

### 2.3. 2D×t Segmentation on Videos

Such methods treat video as a 3D volume and apply CNN structure directly in temporal domain from the input level. It emphasizes the dense frame motion equally as the shape in a full-scale video.

**1D×t profiles**: 1D scanning data alone are insufficient for semantic region classification [62]. Dense temporal resolution can compensate the sparsity of spatial resolution and enhance object detection with motion [42,43]. These works apply semantic segmentation along time axis but involve large data overlap in computing.

**2D×t video volume**: applies 3D-CNN [44-51] and 3D semantic segmentation [52] to encode video spatially and temporally to produce voxel-wise output. We also adopt CNN based pyramid in the temporal domain as the ultimate access of video. It achieves temporal association (continuity and causality) up to T frames. The temporal data are also learned through convolution and maximum pooling, as this has been proved as the most effective way in spatial learning. The temporal changes of objects appear as tubes in video volume dividable 3D regions. However, due to large memory usage and intensive computing, this approach has not been implemented widely.

The 3D segmentation applied to a video clip [52] has latency (patch-mode). For online tasks, a naïve shift of network [34] has large data overlaps in time and can hardly catch a video input rate. To solve this bottleneck for real-time applications, our work derives non-redundant SMN from the naïve shift-mode for both **1D×t** and **2D×t** streaming data, which reuses the computed values when the network shifts over input. The memory of SMN will be much smaller than the default patch-mode. Different from other network variants that add ad-hoc links, our SMN is logically derived from CNN network such that it guarantees the same accuracy as backbone patch-mode on video and naive shift-mode (as our baseline).

## 3. Sequential Semantic Segmentation

### 3.1. Training of temporal network – Patch Mode

The semantic segmentation employs pyramid structure in encoding and decoding modules [10]. Taking an encoder for 2D×t streaming data as an example, we apply the 2×2×2 kernel size with stride of 2 at each dimension for maximum pooling. The feature map will then be reduced to half size layer-by-layer. To down-sample an input video cube from T to 1 in temporal domain, the number of layers is $log$T.

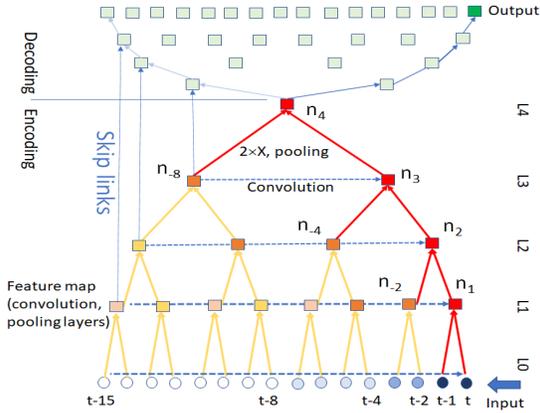

Figure 2: Temporal view of sequential semantic segmentation net (4 layers with input at L0) in shift-mode. Segnet as backbone and spatial domain in pyramid omitted here. Multi-level and multi-span referencing to the past is embedded in the structure. Dash lines: convolutions. Solid lines: maximum pooling from subtrees. Red lines: computing newest input.

When semantic segmentation includes the temporal domain, it can be applied in patch-mode without data overlapping (Fig. 1) for off-line but not for time-critical responding; this is not an image semantic segmentation alone. For real-time decision-making or action planning, a shift-mode on dynamic images is necessary for input frame as in Figure 2. To classify spatial input $I(x,t)$ to a class $C_i$, $i = 1, \ldots, C$ based on past T frames, i.e.,

$$x(t) = y \mid I(x, t), I(x, t-1), \ldots, I(x, t-T+1), y \in C_i \quad (1)$$

Thus, the temporal period T involved in segmentation has overlap between consecutive frames. A memory is increased T time if the history span is large (T has 30 frames in a second of video). It is not affordable for many edge devices and sensors.

In many cases, motion tells more truth when spatial appearance has ambiguity. For example, 1D line from driving video does not reveal sufficient shape in the road environment, but their temporal concatenation in 1D×t road profile (Fig. 3) under a smooth vehicle motion provides more shape of road [62] for path planning, and trajectories of vehicles [42] and pedestrians [43] for collision avoidance. Multiple road profiles can yield a headway for speed planning of a vehicle [57].

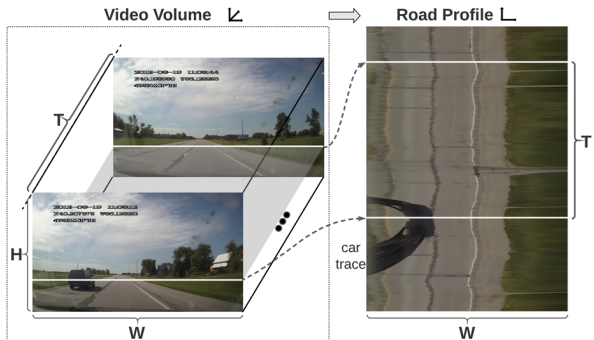

Figure 3: 1D lines scanned in from a camera on ego-vehicle to form a road profile. It contains the road position at a depth, and vehicle positions ahead at every moment.

For 2D videos, many sequences are hard to learn in spatial domain due to less samples on complex scenes (e.g., in DAVIS [54], dancers have articulation motion), different aspects, heavy occlusion, and changes in costume. While motion provides stable flow against background. We train the semantic segmentation network involving T=32 fr. (about one second) to classify video frames.

### 3.2. Testing phase of network – Shift Mode

In the testing phase, the patch-mode needs to wait data streamed in to fill a full patch, which yields a long delay in output after computing the whole patch. It can only be used for off-line data mining rather than online decision making in real-time machine control. Therefore, we apply a shift mode of network to generate output for each input. As shown in Fig. 2, input streams in frame-by-frame. Assume the network looks back T elements to infer the current output. The pyramid structure of the encoding module is implemented by 2×2 (temporal×spatial) filtering locally, and then 2×2 maximin pooling. Their strides are 1 and 2, respectively. The network reaches one latent node through $logT$ layers (ignoring sub-layers). Since the network performs shift-mode and the top node passes every input, setting more than one top nodes in temporal domain only repeat the same operation and are totally unnecessary. The decoding component only needs to output latest frame through green nodes in Fig. 4 with zero-latency.

## 4. Shift-Memory Net without Redundancy

### 4.1. Periods of network subtrees over input data

In shifting the pyramid encoding network (Fig. 2), the upward abstraction is computed between subtrees through node convolution and maximum pooling. The red edges computes subtrees involving the newest input. Because the convolution at each layer uses the same coefficients learned in the training phase, it is noticeable that the subtrees at each level have large overlaps over consecutive input as the network moves a step rightward, i.e., input data streamed in each moment. The old subtrees (in orange) in the network were actually computed by the rightmost subtrees (red nodes) a while ago during shift.

For example, subtree $n_{-2}$ was computed by subtree $n_1$ two steps earlier. Subtree $n_{-8}$ was computed 8 steps earlier by subtree $n_3$ when it was over the same input. According to layer L in the network, this repeat has a period $2^L$, as can be confirmed in Fig. 2. Between the periods, the node values are different during the network shift, as shown by blank boxes in Fig. 4. This redundancy in computing subtrees can be avoided by reusing the values one-period ago at nodes $n_1,...,n_3$, as long as every value at the newest nodes (red in Fig. 4) is saved in the memory.

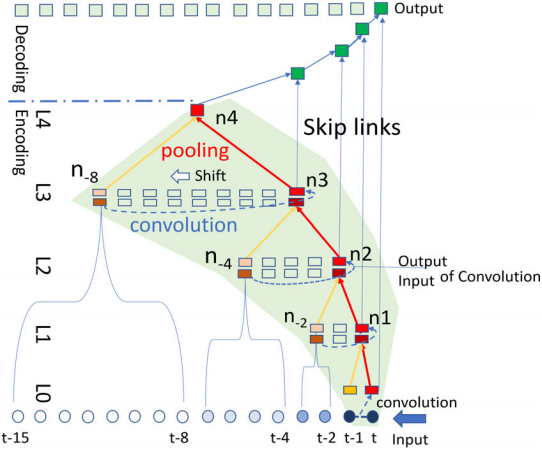

Figure 4: Shift-Memory Network (SMN) by computing the newest nodes (red boxes) with the key nodes separated apart (orange) and saving newly computed nodes in the memory nodes (blank boxes). The green part is the real SMN memory.

### 4.2. SMN derived from shift-mode

We have found a large redundancy in the network computation on overlapped data in naïve shift-mode. Based on the structure shown in Fig. 4, we design a new network for the testing of sequential semantic segmentation. At each level of the pyramid network, time cells of memory are prepared to save the latent variables computed at the very front of the network (red nodes from newest input). The repeating period at each layer is $2^L$, which counts 2, 4, 8 at layer 1, 2, 3 as in Fig. 4. For the newest input, the computation is done upwards to the top, and the variables are saved in the memory nodes for later use. At level L, the convolution is done between the newest node and past key node $2^L$ frames ago, and the result is further pooling with the key node to upper layer. Both values are saved in memory denoted as M(P,C) for later convolution at the same layer and maximum pooling to the upper layer. The total memory at level L has $2^L+1$ nodes including new node at end and $2^L$ previous nodes. They shift one step to the past for next input.

Logically, such memory in blank boxes in Fig. 4 shifts step by step as the network reads input data incrementally. In real practice, we use circular memory without real physical shift. We name this structure Shifting-Memory Network, with network coefficients extracted from the pyramid learned in patch-mode training.

At the starting time, the network ignores output for a period T as the SMN has not been filled up with all valid input. During the first period shift, the subtrees at front end (red) are computed gradually and the values are shifted stepwise backward until nodes in all layers are in position. This non-redundant SMN largely saves computing as compared to shift-mode (baseline) while keeps the equivalent accuracy, because SMN is logically derived from baseline model rather than its modification with additional links.

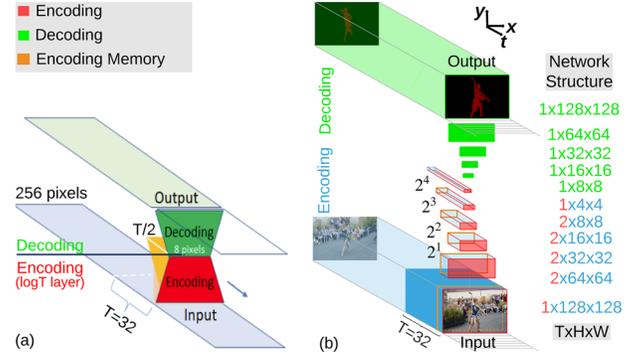

Figure 5: SMNs working on-line input and video respectively. (a) 1D×t SMN, and (b) 2D×t SMN. The memory (orange volume) is computed with new input (red surface) in encoding and the green surface is for decoding process.

### 4.3. SMN for different spatial dimensions

For driving scenes, continuous 1D horizontal input (1D×t) captures surrounding motion directly as trajectory, as vehicles move on flat surface. The horizontal motion flow can provide crucial information for driving. If the input is 1D data in the spatial domain as in Fig. 5(a), the line is down-sampled temporally via 2×2 maximum pooling as the layer increases. According to our derivation above, the shift memory is the volume between the red front surface and the orange back surface in Fig. 5(a). The higher the level, the longer the memory length is, and the shorter the spatial line shrinks. The SMN only computes convolution and maximum pooling at nodes on the front surface (in red) with the past surface (in orange). This saves a vast amount of computing in the shifting pyramid. The complexity is only around the front surface (in red). The decoding produces the latest line through the green surface spatially expended from latent nodes in Fig. 5(a). Such a network works for the road profiles from driving video, line scanning for pedestrian, and LiDAR in surveillance.

For 2D images in the spatial domain, Figure 5(b) shows its SMN in encoding and decoding. The convolution and maximum pooling are then performed on 2×2×2 cube to yield single node. The latent nodes in SMN have more time cells to memorize and longer time spans than those 2D+t networks [34,38] as the layer gets high. The output is the latest frame, and the piled latest frames form video volume.

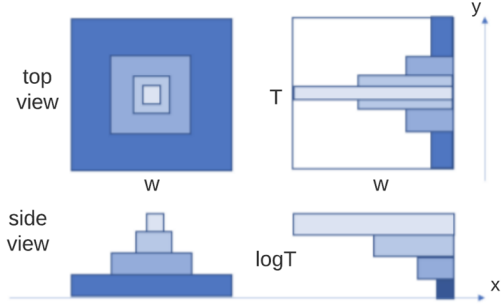

Figure 6: Memory map for traditional CNN and SMN in encoding modules. Light colors indicate high layers in network. Aligning SMN righward, it has the smaller memory but produces the same result as left.

### 4.4. Memory and Computing Complexity of SMN

With the SMN, the memory kept at each moment for 1D lines and 2D frames can be deduced in Table 1, which are calculated as follows. Assume the length of the line is $w$ times of T, which is $2^L$. When we count the memory, we include those spatial nodes/cells as in Fig. 6. If the pyramid patch contains $W \times T$ pixels, where $W=T$ as depicted in Fig. 6. We examine the patch T×T in spatial and temporal domain. The pyramid has the memory size (number of memory nodes) from bottom to top

$$PM_1 = T^2 \Sigma_i (1/4)^i = 4(4^L -1)/3 \approx 4T^2/3 \qquad (2)$$

which is the number of filtering computation.

For the SMN, the non-redundant filtering in the front face is calculated for lines

$$Com_1 = T+T/2+T/4+T/8 \ldots 1 = 1+2+\ldots 2^L = 2T-1. \qquad (3)$$

where the convolution and maximum pooling at a node is counted as a single calculation. For SMN, the total memory nodes is calculated as

$$SM_1 = T+2T/2+4T/4+ \ldots + 2^L = TL = T\log T. \qquad (4)$$

If $T$ is selected as 32 where $L$ is 5, the SMN has a memory of 160 nodes reduced from 1024 nodes in the pyramid.

For a dense 2D image sequence, the pyramid has its nodes in cubes

$$PM_2 = T^3+T^3/8+ T^3/64+ T^3/512+\ldots+1$$
$$= 2^{3L} +2^{3L}/2^3 +2^{3L}/2^6+ \ldots 2^{3L}/2^{3L} = 8(2^{3L} -1)/7 \approx T^3 \qquad (5)$$

For image, the filtering is the same number of convolutions on cube

$$COM_2 = T^2 + T^2/4 + T^2/16 + T^2/64 + T^2/256 \ldots$$
$$= 4T^2 (1-1/4^L)/3 = 4(T^2-1)/3 \approx T^2 \qquad (6)$$

for square convolution.

For image based SMN, the memory from periodicity are

$$SM_2 = \Sigma\, 2^i\, (T/2^i)^2 = \Sigma\, T^2/2^i = T\,(2^L -1) = T^2 \qquad (7)$$

The computation and memory save one dimension as described in Table 1. Overall, the memory of SMN reduced from T in shift-mode of pyramid to $\log T$ times, while performs whole span time referencing as shift-mode. The computation complexity is also dropped one dimension from naïve shift-mode (thus is the same order as patch-mode in off-line data processing).

## 5. Experiments

The experiments are carried out on standard benchmark datasets including 1D driving profiles [42,53], Highway Driving Dataset [56], and DAVIS [54,55], because of their full frame rate annotation (24Hz or 30Hz). We first test 3D-semantic segmentation baseline that is naïve shift-mode trained in patch-mode, and then apply SMN with the same parameters for faster inferencing in a smaller memory. This will confirm our structure design of SMN and its effectiveness for sequential semantic segmentation. The baseline shift-mode further compares with static image segmentation using the same backbone networks, which shows a higher accuracy with motion continuity and causality than that using single image. This is generally true if the spatial training on limited samples still has some space to improve.

### 5.1. Training Patch-mode and Testing Shift-mode

**Data pre-processing**. The training and testing are based on real datasets. The 1D×t video profiles [53] will be horizontally resized to W=256 pixels in width. We use temporal history T=32fr. with shifting-unit Δt=8fr. in patch training. In testing phase, SMN uses Δt=1 in shift-mode to generate output in zero-latency. Thus, each input patch size cut from the road profile is 32×256 pixels (Fig. 2). The backbones used for training are all 2D-CNN networks.

The training for Highway Driving dataset (HWD) [56] and DAVIS [54,55] are on 3D video volumes labeled at every voxel. The testing is on the latest frame to provide segmented regions without latency. The details of preprocessing are given in Table 2.

**Network setting**. Training 1D×t video profiles uses multiple backbone models of image semantic segmentation [10-16]. For training 2D×t in patch-mode, we adapted Segnet [10] and FrrnB [13] to 3D-semantic segmentation baselines (shift-mode) for streaming data. Since the network of semantic segmentation needs to respond to every input frame, we make the top layer of encoding block to be a single array of latent variables to scan temporal data. More than one array implies a repeating work in temporal shift, i.e., a wasting. When T is set as 32, five encoding and decoding layers are designed. At each layer, convolution and maximum pooling uses 2×2×2 kernel size, along with batch normalization and activation function. The stride of convolution is 1, while the stride of pooling is 2. Thus, a video volume of 32×128×128 (T×H×W) is down-sampled to 1×4×4 after encoding and scaled back in decoding module. For backpropagation, we use RMS as optimizer and set a decayed learning rate from 0.0001.

| Dimension | Num. nodes in net | | Computation nodes/frame | | |
|---|---|---|---|---|---|
| | SMN | Patch or Shift PM | SMN | Shift mode | Patch mode |
| 2D (1D×t) | T logT | $4T^2/3$ | 2T-1 | $4T^2/3$ | $T^2$ / T |
| 3D (2D×t) | $T^2$ | $T^3$ | $4T^2/3$ | $T^3$ | $T^3$ / T |

Table 1: Memory and computing for line and image sequences in different modes. The SMN saves one dimension data processing.

| Dataset | 1D×t Profiles clips×frames | HWD video | DAVIS 2017 Video (2D×t) |
|---|---|---|---|
| Training fr. | 30×9k | 15×60 | 60×(25~104) |
| Testing fr. | 10×9k | 5×60 | 30×(25~104) |
| Patch-Size | 32×256 | | 32×128×128 (T×H×W) |

Table 2: Spatial-Temporal data for training and testing. We experiment Video Profiles on both 1D×t and 2D×t datasets.

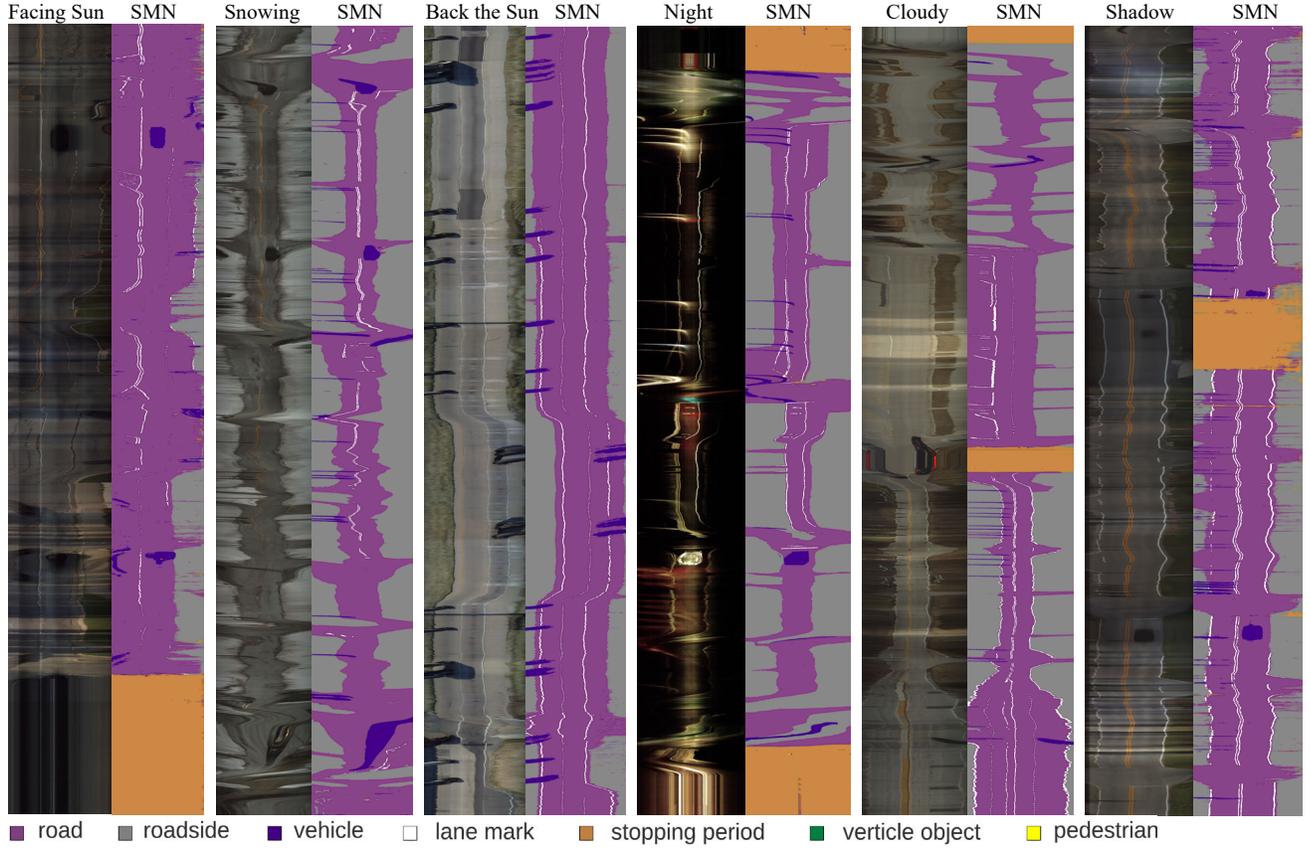

Figure 7: Semantic results of 1D×t video profiles to road and vehicle positions at every moment (backbone training: FrrnB [13], Testing: its SMN). The time axis is upward and each temporal length is 9000 fr. in 5-min driving. The segmented results are robust to illumination changes and the distortion of vehicle traces (inward or outward) due to the relative motion at different positions.

The networks are trained in Tensorflow [58] with a single GPU of NVIDIA RTX 2700 Super but tested without API-based framework. In testing, we write our own programs in part to implement both shift-mode and SMN in order to compare their accuracy and speed fairly. The accuracy is accumulated from all the frames in video clips. The SMN efficiency may not show distinct change as theoretical analysis, when API is employed in part or data size is small. A possible code optimization at API level can convert our substantial data reduction to a real performance time in the future.

## 5.2. Evaluation on Video Profiles

**1D×t Driving Video Profiles** are sampled at horizontal pixel lines from driving video at 30Hz. A temporal-spatial profile I(x,t) contains road layout at the sampled depth and the objects closer than the depth at every moment. There are seven manually annotated semantic class such as road, roadside, lane marks, vehicles, vertical objects, pedestrians and stopping period. These objects, background, and events are inferred from shape and motion in the data stream for path planning and collision avoidance. Compared to normal video frame image, road profile contains more trajectories in the spatially reduced dimension; semantic segmentation is learned more from motion than incomplete shapes. Such driving video clips in 5min have been labeled for training [42], which are much longer than other video sequences. Figure 7 demonstrates the segmented results of SMN for road profiles under various weather illumination conditions, which proves the robustness of segmentation in the spatial-temporal domain, and also displays the motion behavior in a visible way.

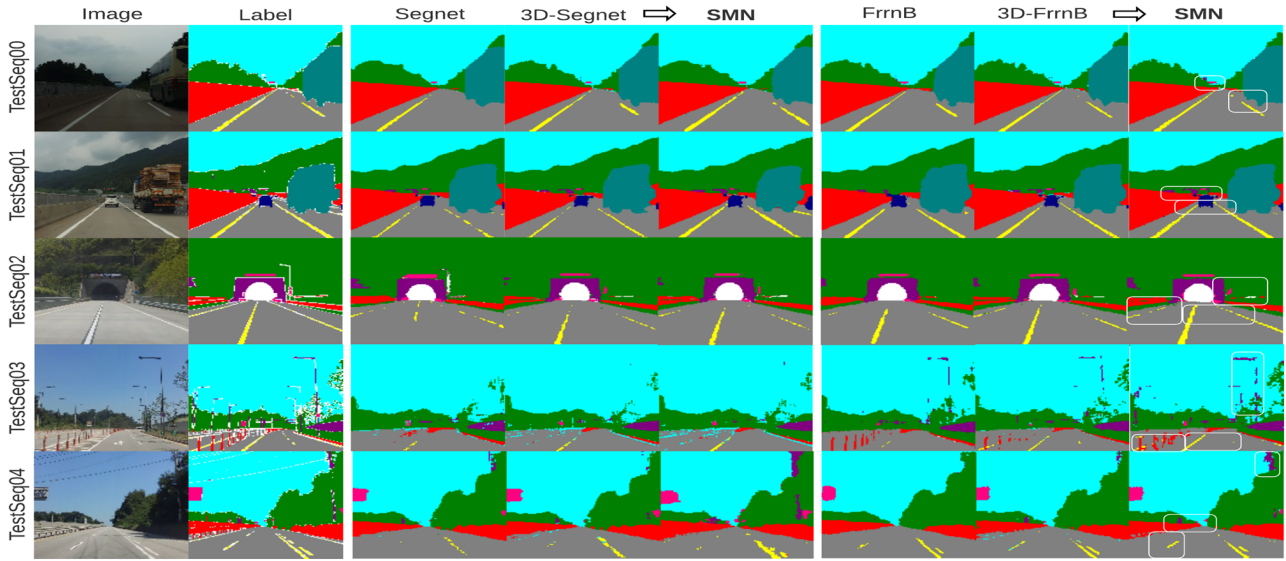

Figure 8: Testing results of SMN on Highway Driving Videos. Compared to 2D image semantic segmentation, baseline shift-mode (3D-SS overtime) enhances the accuracy with zero-latency responding. For 2D×t streaming data, SMN further accelerates the speed, while ensures equivalent accuracy from 3D-SS baseline. Seeing details inside rectangular boxes in last column outperforming other methods. Overall, backbone FrrnB (right block) performs better than Segnet (middle block).

| Backbone | Acc. mIoU (%) | Memory (M) | | Time (ms/line) | |
|---|---|---|---|---|---|
| Base. | Shift & SMN | Shift | SMN | Shift | SMN |
| segnet | 0.83 | 3.4 | 0.2 | 16 | 16 |
| deeplab | 0.88 | 6.8 | 0.7 | 26 | 24 |
| dense | 0.85 | 4.8 | 0.5 | 17 | 17 |
| pspnet | 0.86 | 4.4 | 0.4 | 20 | 20 |
| frrna | 0.90 | 6.9 | 0.7 | 22 | 21 |
| frrnb | 0.92 | 7 | 0.7 | 23 | 23 |

Table 3: Accuracy, run-time memory, and speed for SMN and baseline on 1D×t data from road profiles.

Compared to the baseline shift-mode, SMN only compute the latest line of pixels, and thus it has better efficiency while ensures the equivalent accuracy as shift-mode. Due to the differences in model complexity of backbones (volume of parameters, temp memory, etc.), the testing results varied in time and accuracy. We compare backbone models and SMN in accuracy, memory usage and speed in Table 3. As Tensorflow has not supported the implement of this network architecture, the code based on Tensorflow can not reach the expected time, even if our own program has reached the theoritical values of the time ratio between shift-mode baseline and SMN in Table 1.

### 5.3 Evaluation on Highway Driving Videos

Highway Dataset [56] provides 2-second video clips densely annotated in 30Hz. Since SMN performs in temporal domain, we first compare the accuracy of 3D-shift-mode baselines with image semantic segmentation without time, and then test SMN in zero-latency. In Fig. 8, we highlight results of SMN. Their accuracy in the pixel accuracy (PA) and the mean intersection over union (mIoU) are given for each class [59] in Tables 4, 5.

| Class (Pixel Acc.) | sky | road | lane | fence | construction | traffic sign | car | truck | vegetation | **Average** |
|---|---|---|---|---|---|---|---|---|---|---|
| Segnet w/o time | 0.98 | 0.97 | 0.50 | 0.83 | 0.46 | 0.67 | 0.91 | 0.90 | 0.92 | 0.79 |
| 3D-Segnet | 0.99 | 0.99 | 0.52 | 0.83 | 0.49 | 0.69 | 0.99 | 0.94 | 0.96 | 0.82 |
| **SMN of 3D-Segnet** | 0.99 | 0.98 | 0.54 | 0.88 | 0.56 | 0.70 | 0.99 | 0.93 | 0.95 | 0.83 |
| FrrnB w/o time | 0.99 | 0.99 | 0.71 | 0.93 | 0.57 | 0.71 | 0.93 | 0.92 | 0.98 | 0.87 |
| 3D-FrrnB | 0.99 | 0.99 | 0.76 | 0.89 | 0.63 | 0.81 | 0.99 | 0.96 | 0.98 | 0.89 |
| **SMN of 3D-FrrnB** | 0.99 | 0.99 | 0.79 | 0.90 | 0.69 | 0.83 | 0.99 | 0.95 | 0.98 | **0.90** |

Table 4: Pixel Accuracy of each semantic class in Highway Driving Dataset. Class of unknow is omitted.

| Class (mIoU) | sky | road | lane | fence | construction | traffic sign | car | truck | vegetation | **Average** |
|---|---|---|---|---|---|---|---|---|---|---|
| Segnet w/o time | 0.95 | 0.91 | 0.35 | 0.68 | 0.41 | 0.45 | 0.78 | 0.84 | 0.89 | 0.70 |
| 3D-Segnet | 0.96 | 0.91 | 0.38 | 0.73 | 0.43 | 0.48 | 0.96 | 0.87 | 0.88 | 0.74 |
| **SMN of 3D-Segnet** | 0.95 | 0.92 | 0.42 | 0.76 | 0.44 | 0.56 | 0.97 | 0.91 | 0.89 | 0.76 |
| FrrnB w/o time | 0.97 | 0.93 | 0.57 | 0.82 | 0.48 | 0.63 | 0.79 | 0.87 | 0.9 | 0.79 |
| 3D-FrrnB | 0.97 | 0.93 | 0.62 | 0.82 | 0.55 | 0.72 | 0.98 | 0.93 | 0.92 | 0.83 |
| **SMN of 3D-FrrnB** | 0.97 | 0.94 | 0.65 | 0.81 | 0.54 | 0.71 | 0.98 | 0.92 | 0.91 | **0.83** |

Table 5: Accuracy of mIoU of each semantic class in Highway Driving Dataset. Class of unknow is omitted.

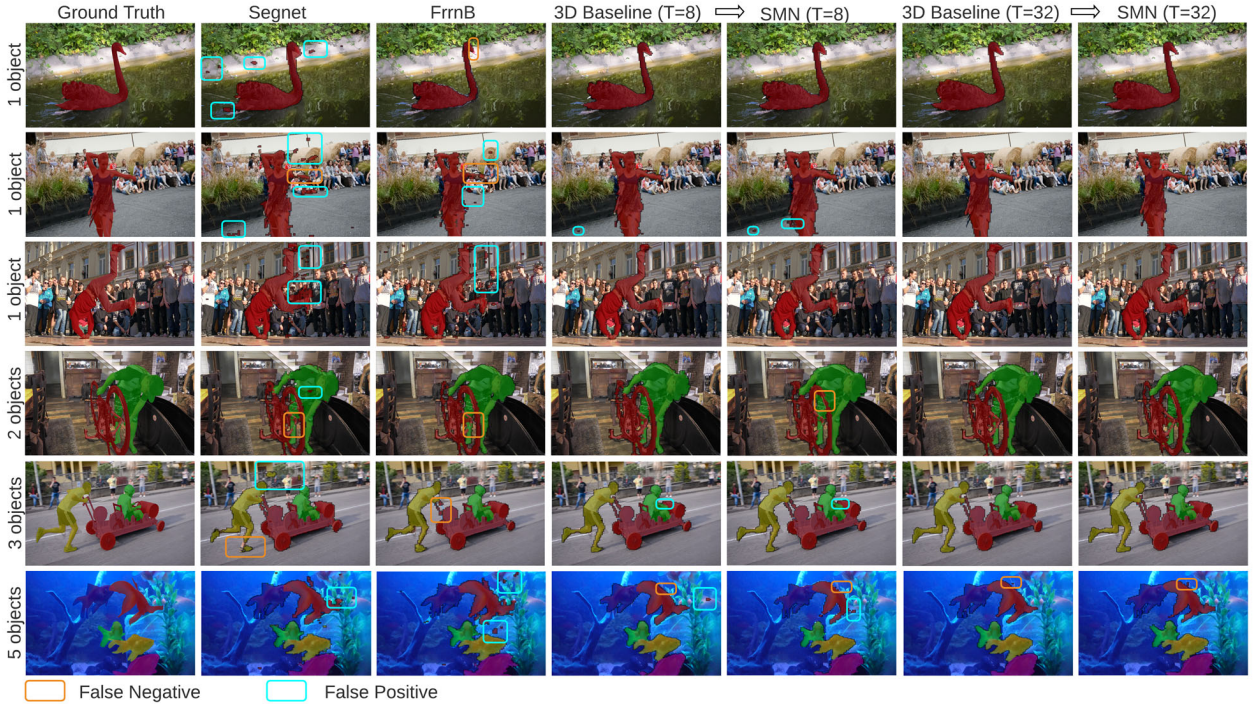

Figure 9: Testing results on DAVIS. Compared to 2D image segmentation (Segnet [10], FrrnB [13]), our 3D video baselines and SMN exploit temporal information that enhanced motion consistency. After training the 3D baseline with T=8 and T=32. The results show that a multi-span at multilevel temporal association has better continuity and less noise (columns 2, 3 versus other columns). SMN further avoids the 3D data overlap in network and the redundant computation in shift-mode for zero-latency output.

Among the methods with and without temporal information, 3D-semantic segmentation on video volume has advantage over image segmentation in accuracy. This confirms that the temporal continuity is suitable for segmenting smooth motion. The problems remained for testing with 3D-semantic segmentation baseline (naïve shift-mode) are large memory and computational redundancy, which are further solved by our SMN in shift-mode. The overall statistical results in Tables 4 and 5 show equivalent accuracy of SMN to its 3D-baseline. The multi-span and multilevel memory in SMN enhanced its power in capturing motion consistently, as highlighted in Fig. 8 at lane and far scenes.

### 5.4 Evaluation on DAVIS Object Segmentation

In addition to driving videos, we apply SMN to 3D-SS on DAVIS mainly for verifying its efficiency and memory usage over baseline, since DAVIS is a motion-orientated dataset. Although DAVIS is non-semantic (with BG and FG only), we report the same metrics as in [54] such as average region similarity (J) and contour accuracy (F) in Table 6. The results are visualized in Fig. 9 as well. We further compare SMN to other methods by plotting them in an accuracy-efficiency trade-off chart displayed in Fig. 10.

SMN employs multi-span and multi-level temporal memory and can improve mIoU by at least 5% compared to 2D image segmentation.

Overall, our method on SMN achieves state-of-the-art performance with experiments. (1) 3D-semantic segmentation on video volumes outperforms single image semantic segmentation, but adds additional memory and computation. (2) Thus, our SMN further reduces data overlap and redundancy of 3D-semantic segmentation in shift-mode. SMN maintains the equivalent accuracy as baseline but uses much smaller memory. It consumes about twice of the computation cost in segmenting a single 2D image. (3) We adjust the temporal length of 3D-semantic segmentation according to motion and action duration for SMN to cope with edge devices without accuracy loss.

|  | Methods | J | F | Memory | Time |
|---|---|---|---|---|---|
| 2D | Segnet | 0.66 | 0.54 | 6.7M | 46.7 ms |
|  | FRRNB | 0.70 | 0.61 | 13.9M | 74.7 ms |
| 2D×t (Segnet) | Baseline T=8 | 0.76 | 0.71 | 36.5 M | 88.2 ms |
|  | SMN (T=8) | 0.75 | 0.70 | **13.1 M** | 63.4 ms |
|  | Baseline T=32 | 0.78 | 0.74 | 145.6 M | 119.3ms |
|  | SMN (T=32) | 0.76 | 0.72 | **15.2 M** | 87.1 ms |

Table 6: Quantitative evaluation on DAVIS17 by Segnet and FRRNB. Video frames are reduced to 128×128 pixels for segmentation networks. All the methods are tested on local machine with a single GPU of NVIDIA RTX 2700 Super.

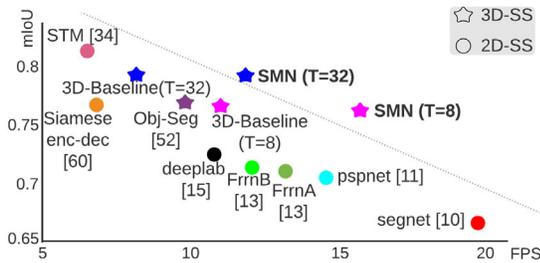

Figure 10: A comparison of quality and speed for semantic segmentation methods on DAVIS-2017 benchmark including 2D and 3D-SS methods, our 3D-SS baselines and SMNs in testing. We visualize mIoU against frame-per-second (FPS).

The SMN is not a network adding more links across layers and time, but reducing redundancy from a full-size volume with the most complete temporal associations. SMN can be converted from most backbones in 2D semantic segmentation, and then from the corresponding 3D segmentation baselines. On the other hand, SMN certainly works for 3D-CNN without a decoding module for object detection.

## 6. Conclusion

With sequential inputs in many computer vision tasks, we extend semantic segmentation to temporal domain utilizing full-scale motion, which increased accuracy compared with image segmentation. To solve the problems of large memory and redundancy in the implementation of CNN-based pyramid, we convert the network to Shif-Memory Network for fast inference. It decomposes the pyramids in shift-mode to remove the redundancy in overlapped data. Trained on large machines or clouds in patch-mode, SMN tests streaming data from linear sensors (such as LiDAR) and videos in a much smaller memory and less computation cost than shift-mode. We confirm its compatible accuracy and higher performance for both 1D and 2D streaming data on benchmark datasets. Our designed network will make semantic segmentation available on edge devices for time-critical applications.

## References

[1] M. Yang, S. Wang, J. Bakita, T. Vu, F. Smith, J. Anderson, J. Frahm, Re-think CNN frameworks for time-sensitive autonomous driving applications: addressing an industrial challenge. *IEEE Real-Time and Embedded Technology and Applications Symposium (RTAS)*, 2019.

[2] S. Ren, K. He, R. Girshick, J. Sun, Faster R-CNN: Towards real-time object detection with region proposal networks. *Proceeding of the 28th NIPS*, 91-99, 2015.

[3] X. Peng, C. Schmid, Multi-region two-stream R-CNN for action detection. *ECCV*, pp. 744–759, 2016.

[4] Z. Cai, N. Vasconcelos. Cascade r-cnn: Developing into high quality object detection. *CVPR*, 6154-6162, 2018.

[5] J. Redmon, S. K. Divvala, R. B. Girshick, A. Farhadi, You Only Look Once: Unified, Real-Time Object Detection. *CVPR*, 2016, pp. 779–788.

[6] J. Remon, A. Farhadi, YOLO9000: Better, Faster, Stronger. *CVPR*, 2017, pp. 617–6525.

[7] J. Choi, D. Chun, H. Kim, H. Lee. Gaussian Yolov3: An Accurate and Fast Object Detector Using Localization Uncertainty for Autonomous Driving. *ICCV*, 2019.

[8] W. Liu, D. Anguelov, D. Erhan, C. Szegedy, S. Reed, C. Fu, and A. Berg. SSD: Single Shot Multibox Detector. In *ECCV*, pp. 21-37, 2016.

[9] J. Long, E. Shelhamer, T. Darrel, Fully convolutional networks for semantic segmentation, *IEEE CVPR*, pp. 3431-3440, 2015.

[10] V. Badrinarayanan, A. Kendall, R. Cipolla, SegNet: A deep convolutional encoder-decoder architecture for image segmentation, *IEEE Trans. Pattern Anal. Mach. Intell.*, 39(12), 2481-2495, Dec. 2017.

[11] H. Zhao, J. Shi, X. Qi, X. Wang, J. Jia, Pyramid scene parsing network, *IEEE CVPR*, 2017.

[12] O. Ronneberger, P. Fisher, T. Brox, U-net: Convolutional networks for biomedical image segmentation, *Inter. Conf. on Medical image computing and computer-assisted intervention*, pp. 234-241, 2015.

[13] T. Pohlen, A. Hermans, M. Mathias, B. Leibe, Fully resolution residual networks for semantic segmentation in street scene, *IEEE CVPR*, 2017.

[14] L. Chen, G. Papandreou, F. Schroff, H. Adam, Rethinking atrous convolution for semantic image segmentation, *CVPR*, 2017.

[15] L. Chen, G. Papandreou, I. Kokkinos, K, Murphy, A. Yuille. Deeplab: Semantic Image Segmentation with Deep Convolutional Nets, Atrous Convolution, and Fully Connected CRFs. *TPAMI*, 2017.

[16] M. Zhen, J. Wang, L. Zhou, T. Fang, L. Quan. Learning fully dense neural networks for image semantic segmentation. *AAAI*, 2019.

[17] P. Isola, J. Zhu, T. Zhou, A. Efros, Image-to-image translation with conditional adversarial networks, *CVPR*, 2017.

[18] T. Yang, M. Collins, Y. Zhu, J. Hwang, T. Liu, X. Zhang, V. Sze, G. Papandreou, L. Chen, DeeperLab: single-shot image parser, In *CVPR*, 2019.

[19] S. Zheng, J. Lu, H. Zhao, X. Zhu, Z. Luo, Y. Wang, Y. Fu, J. Feng, T. Xiang, P. Torr, L. Zhang, Rethinking semantic segmentation from a sequence-to-sequence perspective with transformers, *CVPR*, 2021.

[20] T. Wang, M. Liu, J. Zhu, A. Tao, J. Kautz, B. Catanzaro, High-resolution image synthesis and semantic manipulation with conditional gans, *CVPR*, 2018.

[21] D. E. Rumelhart, G. E. Hinton, R. J. Williams, Learning internal representations by error propagation, *DTIC Document, Tech. Rep.,* 1985.

[22] R. J. Williams, D. Zipser, A learning algorithm for continually running fully recurrent neural networks, *in Neural Computation*, 1989.

[23] S. Hochreiter, J. Schmidhuber, Long short-term memory, *Neural Computation*, vol.9, pp. 1735-1780, 1997.

[24] X. Shi, Z. Chen, H. Wang, D. Yeung, W. Wong, W. Woo. Convolutional lstm network: A machine learning approach for precipitation nowcasting. *In NIPS*, pages 802–810, 2015.

[25] Y. Wang, M. Long, J. Wang, Z. Gao, P. Yu, PredRNN: recurrent neural networks for predictive learning using spatiotemporal lstms, *In NIPS*, 2017.

[26] S. Valipour, M. Siam, M. Jagersand, N. Ray, Recurrent fully convolutional networks for video segmentation. *2017 IEEE Winter Conference on Applications of Computer Vision (WACV)*, pages 29–36, 2017.
[27] S. Nabavi, M. Rochan, Y. Wang, Future semantic segmentation with convolutional lstm, *BMVC*, 2018.
[28] D. Nilsson, C. Sminchisescu, Semantic video segmentation by gated recurrent flow propagation, *CVPR*, 2018.
[29] V. Jampani, R. Gadde, P. Gehler, Video propagation networks, *ICML*, 2017.
[30] E. Yurdakul, Y. Yemez, Semantic segmentation of RGBD videos with recurrent fully convolutional neural networks, *In ICCV*, pp. 367–374, 2017.
[31] J. Donahue, L. Hendricks, M. Rohrbach, S. Venugopalan, S. Guadarrama, K. Saenko, T. Darrell, Long-term Recurrent Convolutional Networks for Visual Recognition and Description, *CVPR*, 2015.
[32] A. Pfeuffer, K. Schulz, K. Dietmayer, Semantic segmentation of video sequences with convolutional LSTMs, *IEEE Intelligent Vehicles Sympo.*, 2019.
[33] A.Vaswani, N. Shazeer, N. Parmar, J. Uszkoreit, L. Jones, A. Gomez, L. Kaiser, Attention is all you need, *NIPS*, 2017.
[34] S. Oh, J. Lee, N. Xu, S. Kim. Video object segmentation using space-time memory networks, In *ICCV*, 2019.
[35] H. Xie, H. Yao, S. Zhou, S. Zhang, W. Sun, Efficient regional memory network for video object segmentation, *CVPR*, 2021.
[36] C. Huynh, A. Tran, K. Luu, M. Hoai, Progressive semantic segmentation, *CVPR*, 2021.
[37] B. Wu, A. Wan, X. Yue, P. Jin, S. Zhao, N. Golmant, A. Gholaminejad, J. Gonzalez, K. Keutzer. Shift: A zero flop, zero parameter alternative to spatial convolutions, *CVPR*, 2018.
[38] J. Lin, C. Gan, S. Han. Tsm: temporal shift module for efficient video understanding, *ICCV*, pp. 7083–7093, 2019.
[39] J. Lin, C. Gan, K. Wang, S. Han, Tsm: temporal shift module for efficient and scalable video understanding on edge devices, *TPAMI*, 2020.
[40] T. Voillemin, H. Wannous, J. Vandeborre. 2D deep video capsule network with temporal shift for action recognition. *In Proceedings of International Conference on Pattern Recognition (ICPR)*, 2020.
[41] Y. Chang, Z. Liu, K. Lee, W. Hsu. Learnable gated temporal shift module for deep video inpainting. *British Machine Vision Conference (BMVC)*, 2019.
[42] G. Cheng, J. Y. Zheng, M. Kilicarslan, Semantic Segmentation of Road Profiles for Efficient Sensing in Autonomous Driving, *IEEE Intelligent Vehicles Sympo*, 564-569, 2019.
[43] G. Cheng, J. Y. Zheng, Semantic Segmentation for Pedestrian Detection from Motion in Temporal Domain, *25th Inter. Conf. Pattern Recognition (ICPR)*, 2020.
[44] D. Tran, L. Bourdev, R. Fergus, L. Torresani, M. Paluri, Learning spatiotemporal features with 3d convolutional networks. *ICCV*, pp. 4489–4497, 2015.
[45] D. Maturana, S. Scherer. Voxnet: A 3D convolutional neural network for real-time object recognition. *IROS*, 2015.
[46] G. Riegler, A. Ulusoy, A. Geiger, Octnet: Learning deep 3d representations at high resolutions. *CVPR*, 2017.
[47] R. Hou, C. Chen, M. Shah, Tube convolutional neural network (T-CNN) for action detection in videos. *ICCV*, pp. 5822–5831, 2017.
[48] R. Hou, C. Chen, R. Sukthankar, M. Shah. An Efficient 3D CNN for Action/Object Segmentation in Video. *British Machine Vision Conference (BMVC)*, 2019.
[49] D. Tran, H. Wang, L. Torresani, J. Ray, Y. LeCun, M. Paluri, A closer look at spatiotemporal convolutions for action recognition. *CVPR*, pp. 6450–6459, 2018.
[50] S. Xie, C. Sun, J. Huang, Z. Tu, K. Murphy, Rethinking spatiotemporal feature learning: 7092 Speed-accuracy trade-offs in video classification. *ECCV*, pp. 305–321, 2018.
[51] M. Zolfaghari, K. Singh, T. Brox. Eco: Efficient convolutional network for online video understanding. *ECCV*, 2018.
[52] L. Sun, K. Jia, D. Yeung, B. Shi. Human action recognition using factorized spatio-temporal convolutional networks. *ICCV*, pp. 4597–4605, 2015.
[53] R. Tian, L. Li, K. Yang, S. Chien, Y. Chen, R. Sherony, Estimation of the vehicle-pedestrian encounter/conflict risk on the road based on TASI 110-car naturalistic driving data collection. *IEEE IV 2014*, 623-629.
[54] F. Perazzi, J. Pont-Tuset, B. McWilliams, L. Van Gool, M. Gross, A. Sorkine-Hornung. A benchmark dataset and evaluation methodology for video object segmentation. *In CVPR*, 2016.
[55] J. Tuset, F. Perazzi, S. Caelles, P. Arbelaez, A. Hornung, L. Gool. The 2017 davis challenge on video object segmentation. *arXiv:1704.00675*, 2017.
[56] B. Kim, J. Yim, J. Kim, Highway driving dataset for semantic video segmentation, *BMVC*, 2018.
[57] Z. Wang, G. Cheng, J. Y. Zheng, Planning Autonomous Driving with Compact Road Profiles. *IEEE ITSC*, 2021, 899-906.
[58] M. Abadi, A. Agarwal, P. Barham, E. Brevdo. TensorFlow: Large-scale machine learning on heterogeneous systems. *Software available from tensorflow.org*, 2015.
[59] M. Berman, A. R. Triki, M. B. Blaschko, The Lovasz-Softmax Loss: A Tractable Surrogate for the Optimization of the Intersection-Over-Union Measure in Neural Networks. *IEEE CVPR*, 2018.
[60] S. Oh, J. See, K. Sunkavalli, S. Kim, Fast Video Object Segmentation by Reference-Guided Mask Propagation. *CVPR*, 2018.
[61] H. Song, W. Wang, S. Zhao, J. Shen, K. Lam, Pyramid dilated deeper convLSTM for video salient object detection. *ECCV*, 2018.
[62] Z. Wang, G. Cheng, J. Zheng, Road edge detection in all weather and illumination via driving video mining. *IEEE Trans. Intelligent Vehicles*, 4(2), 232-243, June 2019.
[63] https://en.wikipedia.org/wiki/Braking_distance
[64] Z. Liu, X. Li, P. Luo, C. Loy, X. Tang, Deep learning markov random field for semantic segmentation. IEEE TPAMI, 40(8), 1814-1828, 2018.
[65] C. Yu, J. Wang, C. Peng, C. Gao, . Yu, N. Sang, BiSeNet: bilateral segmentation network for real-time semantic segmentation. ECCV 2018.
[66] M.Fan, S. Lai, J. Huang, X. Wei, Z. Chai, J. Luo, X. Wei, Rethinking BiSeNet for real-time semantic segmentation. CVPR 2021.